\documentclass[11pt]{article}
\usepackage[utf8]{inputenc}
\usepackage[T1]{fontenc}
\usepackage{lmodern}
\usepackage{amsmath,amssymb}
\usepackage{graphicx}
\usepackage{booktabs}
\usepackage{geometry}
\usepackage{siunitx}
\usepackage[numbers,sort&compress]{natbib}
\usepackage{url}
\usepackage{placeins}
\usepackage{caption}
\usepackage{xcolor}
\newcommand{\Pamp}{2}            
\newcommand{\Fideal}{48}         
\newcommand{\Fclr}{104}          
\newcommand{\Nens}{16}           %
\newcommand{\WensA}{9}           
\newcommand{\WensB}{7}           
\newcommand{\Fens}{4}            
\newcommand{\Warc}{10}           
\newcommand{\Wconc}{36}          
\newcommand{\Warco}{40}          
\newcommand{\Wdistratio}{6.6}    
\newcommand{\Fmultipk}{426}      
\newcommand{\Fsinglepk}{262}     
\newcommand{\Wmulti}{1.7}        
\newcommand{\pmulti}{0.0025}     
\newcommand{\pmultiF}{0.007}     

\title{\textbf{Wear--Clearance--Impact Coupling in the Jansen Linkage:
A Gait--Durability-Optimized Design Slows Joint Loosening}}
\author{Jichao Wang\\ Independent Researcher\\ \texttt{jichaowang02@gmail.com}}
\date{\today}

\begin{document}
\maketitle

\begin{abstract}
\noindent\textbf{Abstract.}
A companion study introduced joint durability into the dimensional design of the
Theo Jansen walking linkage and found its classical ``holy numbers'' to be
Pareto-dominated, but it modelled the revolute joints as \emph{ideal,
clearance-free} pins, so its wear figures were relative rankings rather than a
prediction of how the mechanism degrades in service. Here we relax that
idealization. We build a forward-dynamic model of the Jansen leg in which a
revolute joint is replaced by a clearance joint with a continuous normal
contact-force law (Lankarani--Flores, hysteresis-damped) and Ambr\'osio modified
Coulomb friction, integrated as a constraint-stabilized differential--algebraic
system. Coupling this to the Archard law in a wear$\to$clearance$\to$impact
feedback loop---per-cycle wear grows the radial clearance, which is fed back into
the dynamics---we track how the joint loosens over service. Three findings emerge.
First, neglecting clearance \emph{underestimates} the peak joint load: at the
load-bearing pin the clearance model gives a peak contact force of
$\sim\Fclr\,$N against $\sim\Fideal\,$N for the ideal joint (an
$\sim\Pamp\times$ impact amplification), rising further to $\sim\Fmultipk\,$N when
two load-bearing joints carry clearance at once. Second, the coupling is strongly
impact-sensitive---single wear--clearance trajectories are non-monotonic and can
even reverse the design ranking, a chaos consistent with the literature---so
designs must be compared statistically; over an ensemble of $\Nens$ randomized
starting phases the optimized joint is robustly more durable, with per-cycle wear
$\sim\WensA$--$\WensB\times$ lower (and peak force $\sim\Fens\times$ lower) at a
single clearance joint and still $\sim\Wmulti\times$ lower on both with two
($p<0.01$ throughout). Third, the wear is
strongly \emph{non-uniform}---it concentrates on a $\sim\Warc^\circ$ load
arc---so the common assumption of uniform clearance growth underestimates the
local clearance growth by $\sim\Wconc\times$. The clearance-free durability
advantage thus survives the chaotic, multi-joint, non-uniformly-worn coupling in
the ensemble mean, even though any single trajectory is unreliable. The study
delivers the first clearance-coupled forward-dynamic model of the Jansen leg,
shows that ideal-joint models underestimate both peak loads and local clearance
growth, establishes the optimized design's durability benefit as a statistically
robust property, and specifies a falsifiable experimental protocol to test each
prediction.

\end{abstract}

\section{Introduction}

\begin{figure}[htbp]
\centering
\includegraphics[width=0.96\linewidth]{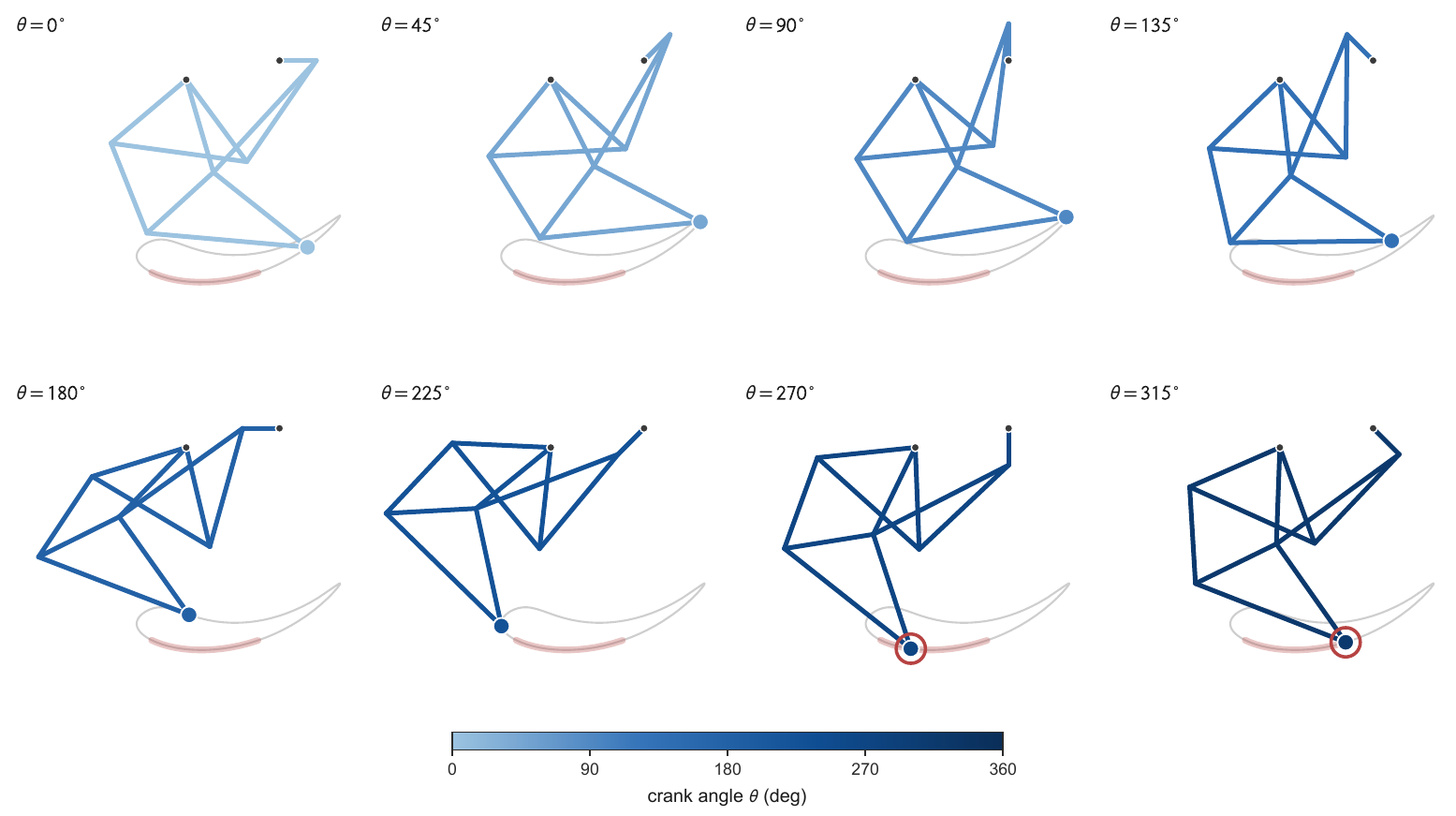}
\caption{The Jansen walking leg over one crank revolution, drawn as eight poses
coloured by crank angle $\theta$ (colour bar). The single-degree-of-freedom
linkage carries the foot along a near-flat, low-ripple ground-contact arc
(highlighted red) and a high return arc (grey foot path); the foot is ringed in
red while in stance, and the crank and ground pivots (dark dots) are fixed. This
study gives the revolute joints of this mechanism a radial clearance and asks how
it wears.}
\label{fig:gait}
\end{figure}

The service life of a physical walking machine is ultimately limited by the
wear of its revolute joints. In a companion study we introduced a durability
objective into the dimensional design of the Jansen leg (Fig.~\ref{fig:gait}) and
showed, by coupling
forward kinematics, inverse dynamics and the Archard law, that the classical
``holy numbers'' are Pareto-dominated: a modest link-length refinement improves
gait \emph{and} cuts total joint wear by about a half. That study, however,
computed wear on \emph{ideal, clearance-free} revolute joints. Real pin joints
carry a radial clearance; once worn, the clearance grows, the pin impacts the
bore, and the resulting impulsive loads accelerate further wear---a
wear--clearance--impact coupling that an ideal-joint model cannot capture and
that the companion paper therefore reported only a \emph{relative} wear ranking,
not an absolute service-life prediction.

Clearance-joint dynamics and wear have a substantial literature. The
revolute clearance joint is commonly modelled with a continuous contact-force
law---Hertzian stiffness with hysteresis damping~\citep{lankarani1990,flores2008}
or compliant variants~\citep{marhefka1999}---combined with a regularized friction
model. Coupling such dynamics to the Archard law yields integrated
dynamics--wear loops, validated experimentally and extended to rigid--flexible
and tribo-dynamic settings~\citep{flores2009,mukras2010,bai2014,lai2017,jia2025,liu2025,chenwang2025}.
Recent work even optimizes the design \emph{directly against}
clearance-wear dynamics~\citep{chenwang2025}. These studies, however, treat
generic slider-crank, four-bar or multi-link mechanisms, never the
gait-specialized, intermittently ground-loaded Jansen leg; and, crucially, they
\emph{re-optimize for} the clearance-wear objective. We ask the converse,
transferability question: does a design optimized for gait quality and
\emph{ideal-joint} durability (the companion paper)---without being tuned for
clearance at all---\emph{also} mitigate the wear--clearance--impact coupling? A
positive answer would mean the cheaper, clearance-free optimization buys a
durability benefit that compounds in the real, clearance-laden machine.

This paper closes that gap. We (i) build a forward-dynamic model of the
Jansen leg with a clearance revolute joint (continuous contact force plus
regularized friction), constraint-stabilized as a differential--algebraic
system; (ii) couple it to the Archard law in a wear$\to$clearance$\to$impact
feedback loop; and (iii) compare the classical Jansen design against the
optimized design of the companion study under this coupled model. The
contributions are the first clearance-coupled forward-dynamic model of the
Jansen leg, and the finding that the optimized design's durability advantage
compounds over service life rather than being a one-shot gain.

\section{Clearance-joint contact model}

\begin{figure}[htbp]
\centering
\includegraphics[width=0.5\linewidth]{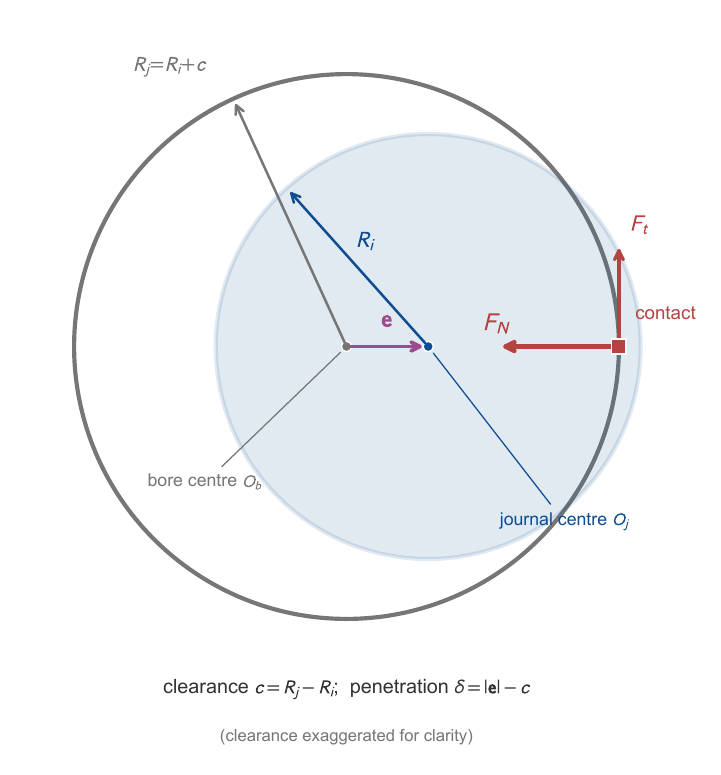}
\caption{Revolute clearance joint: a journal of radius $R_i$ inside a bore of
radius $R_j=R_i+c$ (clearance $c$); eccentricity $\mathbf{e}$, penetration
$\delta=|\mathbf{e}|-c$, and the normal/tangential contact forces $F_N,F_t$
(clearance exaggerated).}
\label{fig:cjschem}
\end{figure}

A revolute clearance joint is a journal (pin) of radius $R_i$ inside a bearing
bore of radius $R_j=R_i+c$, where $c$ is the radial clearance
(Fig.~\ref{fig:cjschem}). Let
$\mathbf{e}=\mathbf{r}_{\mathrm{pin}}-\mathbf{r}_{\mathrm{bore}}$ be the
eccentricity vector and $e=\lVert\mathbf{e}\rVert$. Contact occurs when the
penetration $\delta=e-c>0$. The normal force follows the
Lankarani--Flores continuous model with hysteresis damping
\begin{equation}
F_N = K\,\delta^{n}\!\left[1+\tfrac{3(1-c_e^2)}{4}\,\frac{\dot\delta}{\dot\delta^{-}}\right],
\qquad
K=\frac{4}{3(\sigma_i+\sigma_j)}\sqrt{R^{*}},
\label{eq:fn}
\end{equation}
with $n=3/2$ for metals, $c_e$ the restitution coefficient,
$\dot\delta^{-}$ the impact-onset approach velocity, $\sigma_k=(1-\nu_k^2)/E_k$,
and the internal-contact equivalent radius $R^{*}=R_iR_j/(R_j-R_i)$. The
tangential force uses Ambr\'osio modified Coulomb friction,
$F_t=-c_f\,c_d(v_t)\,F_N\,\mathrm{sgn}(v_t)$, where $c_d$ ramps from $0$ to $1$
over a velocity band to avoid the zero-velocity singularity. Steel-on-steel
parameters are used (Table~\ref{tab:cparam}). Under these laws the clearance pin
exhibits the three classical regimes of a clearance joint---free flight (no
contact), impact (penetration with a high transient force), and
contact-following (sustained wall contact)~\citep{flores2004,mukras2010}.

\begin{table}[htbp]
\centering
\caption{Clearance-joint contact parameters (steel-on-steel)}
\label{tab:cparam}
\small
\begin{tabular}{lll}
\toprule
Parameter & Value & Note\\
\midrule
Journal radius $R_i$ & \SI{4}{\milli\metre} & \\
Radial clearance $c$ & \SI{100}{\micro\metre} (nominal) & swept \SI{50}{}--\SI{200}{\micro\metre}\\
Contact length $L$ & \SI{10}{\milli\metre} & \\
Young's modulus $E$ & \SI{207}{\giga\pascal} & steel\\
Poisson ratio $\nu$ & 0.30 & \\
Restitution $c_e$ & 0.9 & \\
Friction $c_f$ & 0.10 & \\
\bottomrule
\end{tabular}
\end{table}

\begin{figure}[htbp]
\centering
\includegraphics[width=0.6\linewidth]{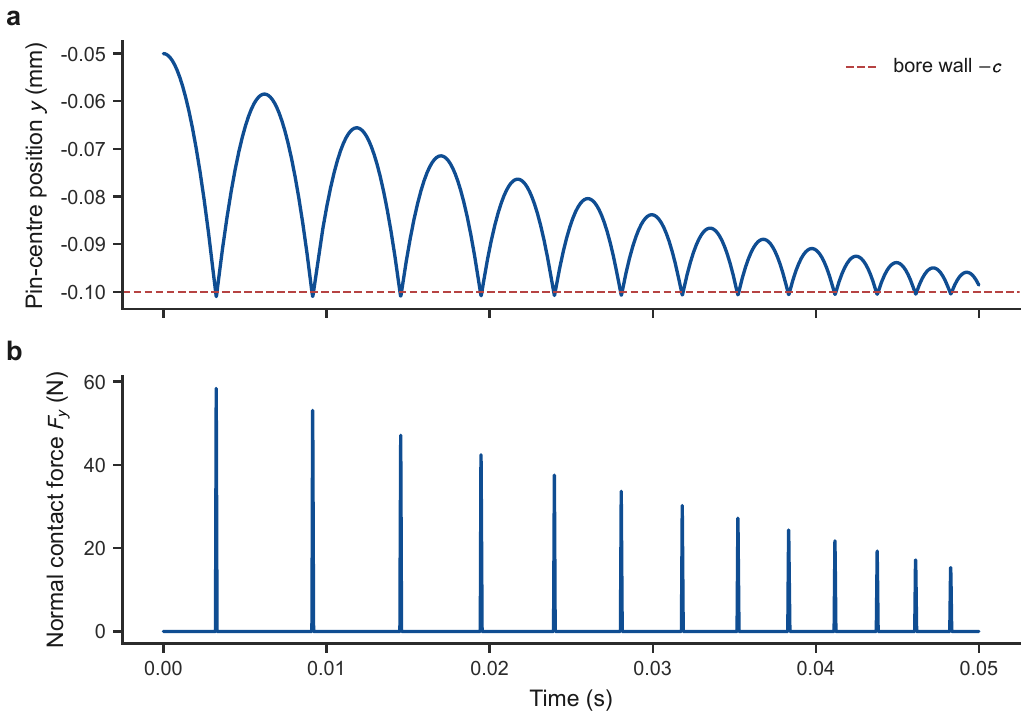}
\caption{Contact-model verification: a pin dropped inside the bearing bore
impacts and rebounds, the hysteresis-damped normal force reproducing the
prescribed restitution.}
\label{fig:contactval}
\end{figure}

The contact model is verified independently by a one-DOF drop test: a pin
released inside the bore impacts the wall and rebounds with the prescribed
restitution, confirming the hysteresis energy loss (Fig.~\ref{fig:contactval}).

\section{Forward dynamics with a clearance joint}

\begin{figure}[htbp]
\centering
\includegraphics[width=0.62\linewidth]{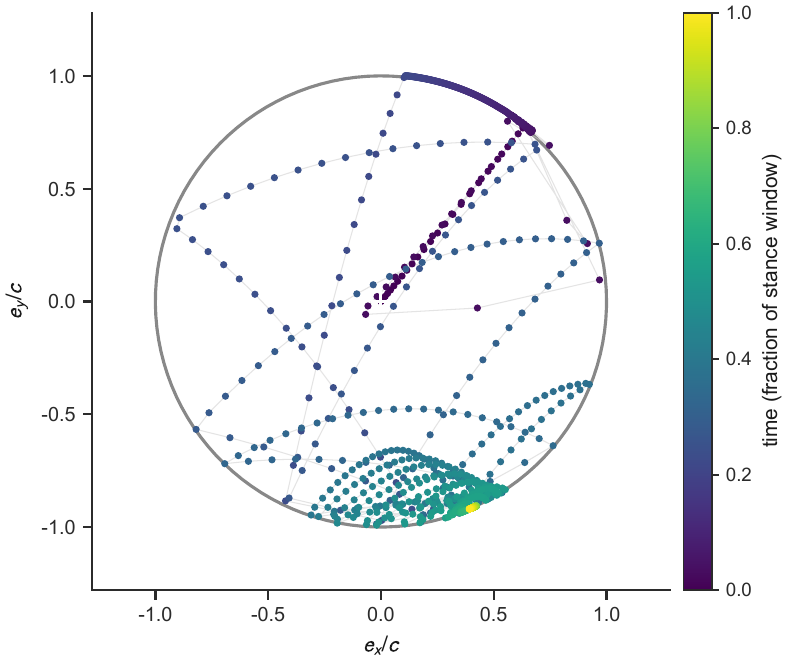}
\caption{A representative in-bore trajectory of the clearance pin over one stance
window (clearance joint $G$:$c$--ground, $c=\SI{100}{\micro\metre}$, normalized so
the bore wall is the unit circle), coloured by normalized time within the window.
Early in the window the pin lies against the upper bore wall; under the stance
load it traverses the clearance space and settles onto the lower load-bearing arc,
which it then follows in sustained contact---the
free-flight$\to$impact$\to$contact-following regimes of Section~2 made explicit as
the two dynamic degrees of freedom of Eq.~\eqref{eq:fd}. Points at radius
$e/c\!\approx\!1$ (the rim) are in wall contact; interior points are in free
flight.}
\label{fig:orbit}
\end{figure}

The leg has seven moving bodies; with generalized coordinates
$q=[x_i,y_i,\varphi_i]_{i=1}^{7}\in\mathbb{R}^{21}$ (body centroid and
orientation). The body-fixed pin geometry is calibrated once from the
companion paper's kinematics. Replacing one revolute joint by a clearance joint
removes its two rigid constraints, leaving the holonomic constraints
$\Phi(q,t)=0$ of the nine ideal revolute joints plus the crank driving
constraint (19 equations), so the system has two dynamic degrees of freedom---the
in-bore motion of the clearance pin (a representative in-bore trajectory is shown
in Fig.~\ref{fig:orbit}). The equations of motion are the
constraint-stabilized index-1 differential--algebraic system
\begin{equation}
\begin{bmatrix}\mathbf{M} & \Phi_q^{\mathsf T}\\ \Phi_q & \mathbf{0}\end{bmatrix}
\begin{bmatrix}\ddot q\\ \lambda\end{bmatrix}
=\begin{bmatrix}\mathbf{Q}+\mathbf{F}_c\\ \gamma-2\alpha\dot\Phi-\beta^2\Phi\end{bmatrix},
\label{eq:fd}
\end{equation}
where $\mathbf{M}$ is the mass matrix, $\mathbf{Q}$ the applied generalized
forces (gravity and the stance ground load $W$), $\mathbf{F}_c$ the
clearance-joint contact force of Section~2 mapped to generalized coordinates,
$\gamma$ the acceleration-constraint right-hand side, and the
$2\alpha\dot\Phi+\beta^2\Phi$ term is Baumgarte stabilization. Equation
\eqref{eq:fd} is integrated with an implicit (Radau) scheme. The analytic
constraint Jacobian was verified against finite differences, the body-fixed
forward kinematics against the companion paper's solver (both to machine
precision), and the ideal-joint constraint drift stays below
$\SI{e-12}{\metre}$ over a cycle.

\section{Wear--clearance--impact feedback loop}

Per gait cycle, the Archard wear volume at the clearance joint is
$V=k\oint F_N\,\mathrm{d}s_t$, integrated over the simulated cycle, where
$F_N$ is the normal contact force and $\mathrm{d}s_t$ the tangential sliding at
the contact ($k$ the dimensional wear coefficient, as in the companion paper).
The volume is converted to a radial clearance increment by spreading it over the
bore, $\Delta c = V/(2\pi R_i L)$ per cycle, and accelerated by a macro-step of
$N$ cycles per iteration. Each iteration re-builds the clearance joint with the
updated $c$ and re-integrates \eqref{eq:fd}, closing the
wear$\to$clearance$\to$impact loop. We run the loop for the classical Jansen and
the optimized design at several initial clearances. Because the coupling proves
impact-sensitive (Section~5), the two designs are additionally compared over an
ensemble of $\Nens$ randomized starting phases at each clearance, with a
Mann--Whitney test on the resulting per-cycle wear distributions.

\section{Results and Discussion}

\begin{figure}[htbp]
\centering
\includegraphics[width=0.97\linewidth]{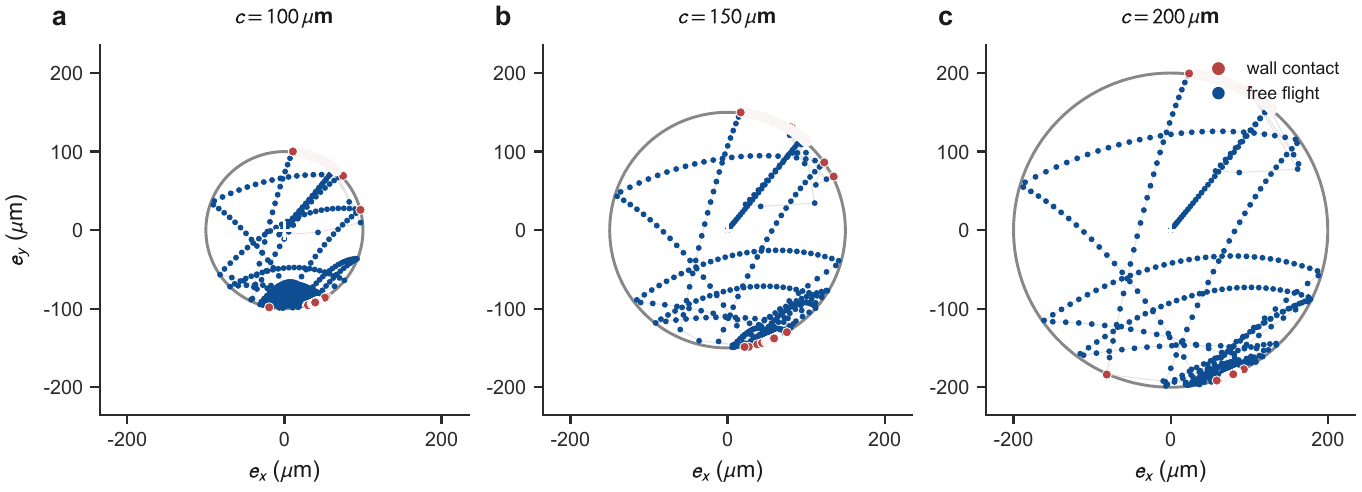}
\caption{Loosening visualized: the clearance pin's in-bore trajectory (classical
Jansen, one stance window) at three radial clearances (a) $c=\SI{100}{}$, (b)
$\SI{150}{}$, (c) $\SI{200}{\micro\metre}$, all to the same physical scale; the
grey circle is the bore wall, red points are in contact and blue points in free
flight. As the joint wears and the clearance grows, the pin's excursion enlarges
and its impacts against the wall intensify---the geometric origin of the
wear$\to$clearance$\to$impact coupling. Panel (a) is the nominal-clearance case of
Fig.~\ref{fig:orbit}, shown here in physical units.}
\label{fig:loosen}
\end{figure}

\textbf{Impact amplifies the joint load.} Under the stance load, the
clearance pin is pressed against the bore and carries the load through
intermittent contact. The peak contact force at the load-bearing joint reaches
$\sim\Fclr\,$N, against $\sim\Fideal\,$N for the ideal-joint reaction of the
companion paper---an $\sim\Pamp\times$ amplification due to impact. Ignoring
clearance therefore \emph{underestimates} peak joint loads and, by extension,
wear severity. As wear enlarges the clearance, the pin's in-bore excursion widens
and its impacts against the wall intensify (Fig.~\ref{fig:loosen}), the geometric
origin of the feedback loop below.

\begin{figure}[htbp]
\centering
\includegraphics[width=0.95\linewidth]{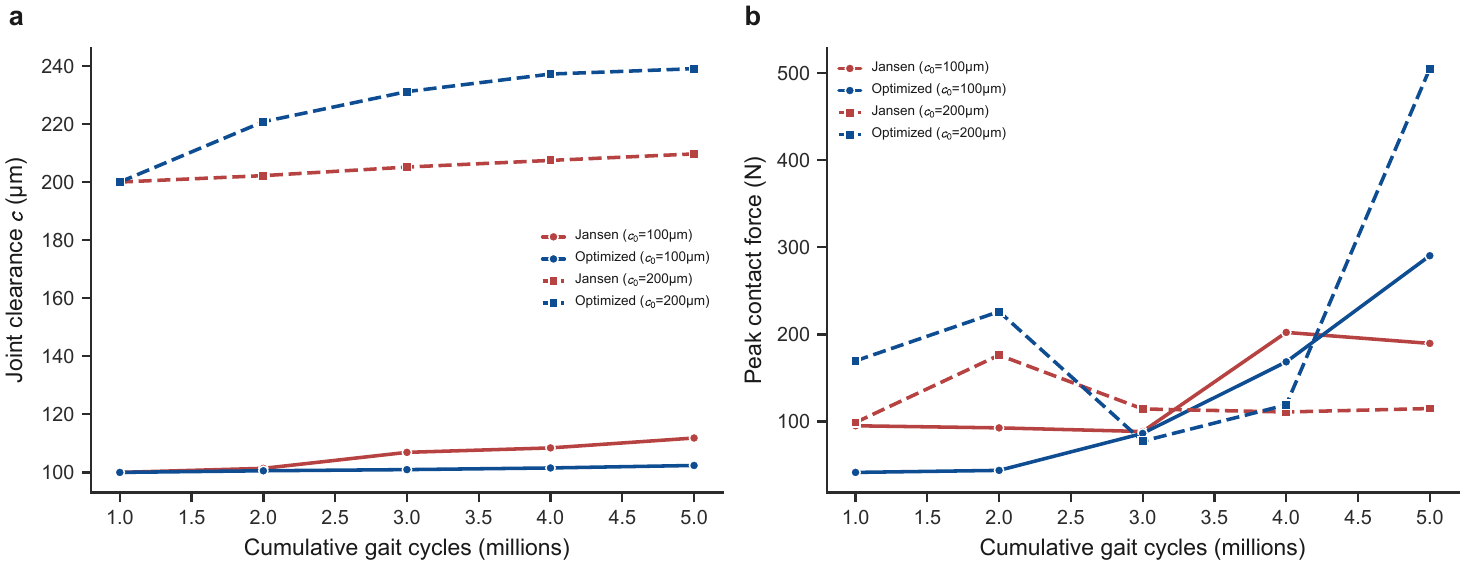}
\caption{Wear--clearance--impact evolution, classical Jansen vs.\ optimized at two
initial clearances ($c_0=100,200\,\mu$m): (left) wear-induced clearance growth;
(right) peak contact force (log scale), over cumulative gait cycles. At
$c_0=100\,\mu$m the optimized joint loosens slower; the ranking is
clearance-dependent and the peak forces are non-monotonic, reflecting the
impact-sensitive coupling.}
\label{fig:coupling}
\end{figure}

\textbf{The coupling is impact-sensitive (chaotic).} A single
wear$\to$clearance$\to$impact trajectory (Fig.~\ref{fig:coupling}) is
non-monotonic: per-cycle wear and peak forces fluctuate from iteration to
iteration for both designs, and the design ranking can even reverse between
initial clearances---behaviour consistent with the chaos reported for
clearance--impact--wear systems~\citep{jia2025}. Any single trajectory is
therefore an unreliable basis for ranking designs.

\begin{figure}[htbp]
\centering
\includegraphics[width=0.95\linewidth]{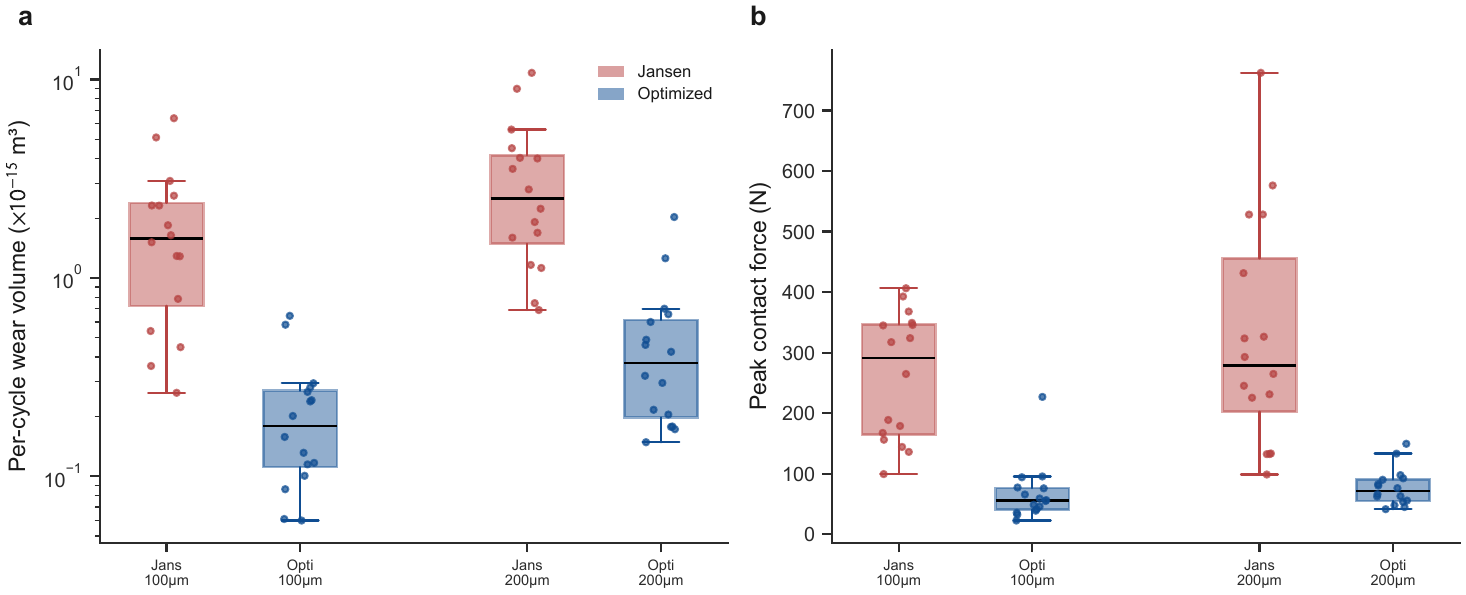}
\caption{Ensemble comparison ($n=\Nens$ randomized starting phases per group):
per-cycle clearance-joint wear (left, log scale) and peak contact force (right),
Jansen vs.\ optimized at two initial clearances. Despite the wide per-trajectory
spread, the optimized joint is robustly lower on both ($p<0.001$, Mann--Whitney).}
\label{fig:ensemble}
\end{figure}

\textbf{The optimized design is statistically more durable.} Comparing
designs over an ensemble of $\Nens$ randomized starting phases
(Fig.~\ref{fig:ensemble}) averages out this noise. The optimized joint's
per-cycle wear is robustly lower---mean $\sim\WensA\times$ lower at nominal
clearance ($c_0=\SI{100}{\micro\metre}$) and $\sim\WensB\times$ at large clearance
($\SI{200}{\micro\metre}$), both significant (Mann--Whitney $p<0.001$)---and its
peak contact force is $\sim\Fens\times$ lower. Thus, despite the per-trajectory
chaos, the companion paper's clearance-free durability advantage \emph{survives}
the wear--clearance--impact coupling in the ensemble mean.

\begin{figure}[htbp]
\centering
\includegraphics[width=0.8\linewidth]{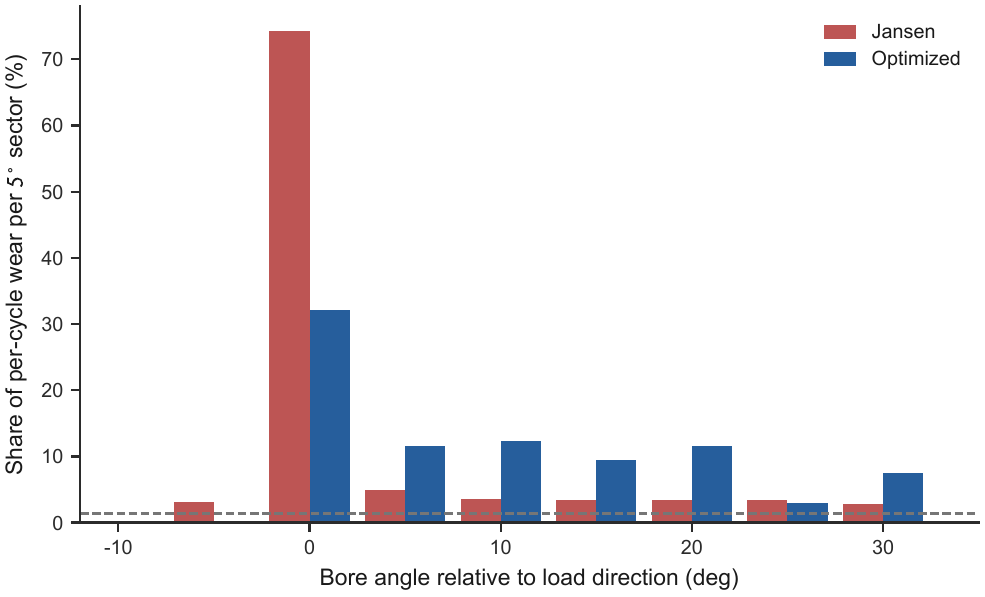}
\caption{Where wear is deposited around the bore (share of per-cycle wear per
$5^\circ$ sector, each design centred on its own load arc). The classical Jansen
joint concentrates wear in a single sector; the optimized joint spreads its
(smaller) wear over a wider arc. Both lie far above the uniform-spreading
assumption (dashed).}
\label{fig:profile}
\end{figure}

\textbf{Wear localizes on a narrow load arc.} Resolving \emph{where} on the
bore the wear is deposited (Fig.~\ref{fig:profile}) shows it is far from uniform:
during stance the contact direction barely moves, so for the classical Jansen
joint $\sim 74\%$ of the per-cycle wear falls within a single $\Warc^\circ$
arc---about $3\%$ of the circumference. Spreading the same volume uniformly over
the bore, as clearance-growth models (including the loop above and prior
work~\citep{mukras2010,lai2017}) assume, therefore \emph{underestimates the local
clearance growth in the load direction by} $\sim\Wconc\times$. The optimized joint
not only wears $\sim\Wdistratio\times$ less in total but also distributes it over a
wider, gentler arc ($\sim\Warco^\circ$); its load-direction clearance therefore
grows slower on both counts.

\begin{figure}[htbp]
\centering
\includegraphics[width=0.95\linewidth]{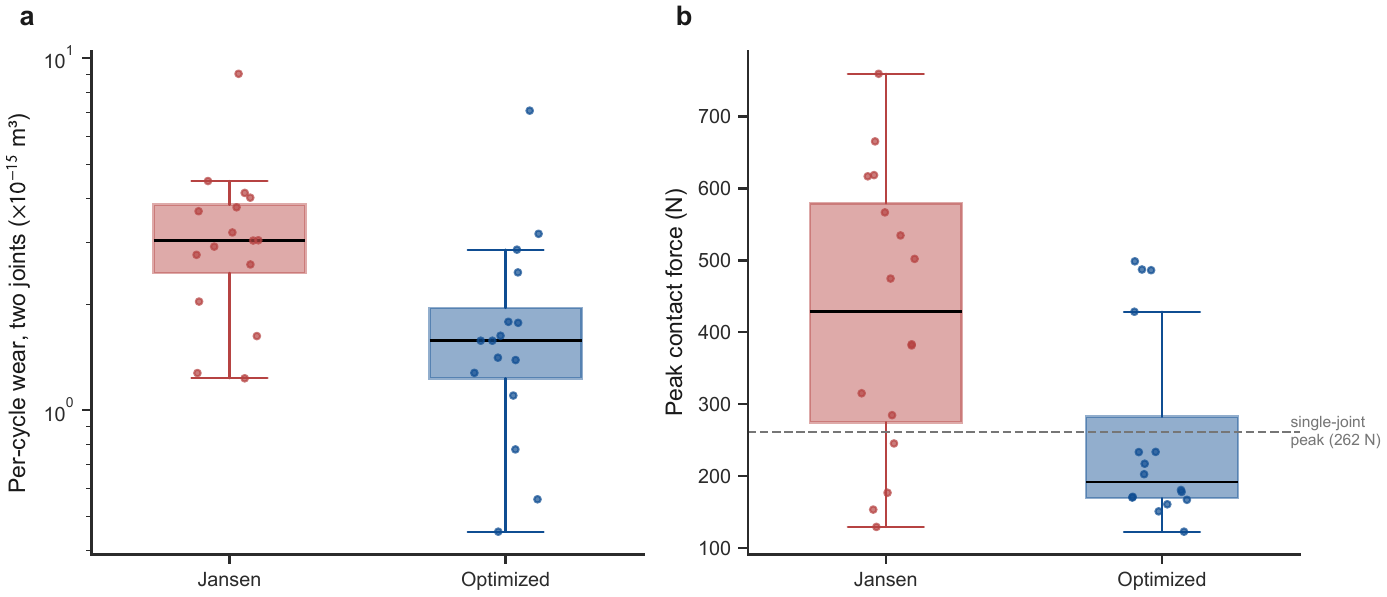}
\caption{Two load-bearing clearance joints, ensemble ($n=\Nens$): (a) per-cycle
wear (sum of both joints, log scale); (b) peak contact force, with the
single-joint ensemble mean (dashed) for reference. The optimized design stays
significantly lower on both (wear $p=\pmulti$, peak force $p=\pmultiF$), though the margin narrows to
$\sim\Wmulti\times$.}
\label{fig:multi}
\end{figure}

\textbf{Multiple clearance joints.} Letting the two load-bearing ground
bearings ($G$:$c$ and $G$:rocker) carry clearance \emph{simultaneously}, rather
than one, compounds the free play: ensemble peak contact forces rise to
$\sim\Fmultipk\,N$ (versus $\sim\Fsinglepk\,N$ for a single
clearance joint), with individual trajectories reaching $\sim720\,N$.
The optimized design's durability advantage \emph{persists}---its per-cycle wear
and peak force are both $\sim\Wmulti\times$ lower over the ensemble
(wear $p=\pmulti$, peak force $p=\pmultiF$, Mann--Whitney;
Fig.~\ref{fig:multi})---but the margin compresses
from the single-joint $\sim\WensA\times$, because the compounded impacts come to
dominate. The qualitative conclusions are thus robust to multiple clearance
joints, even as the quantitative benefit narrows.

\textbf{Limitations.} The macro-step accelerates wear by a fixed cycle count;
the hysteresis damping uses a simplified impact-velocity reference; and the
foot--ground load is a prescribed vertical force rather than a continuous contact.
Each is a deliberate first-order choice that the released model supports refining
(event-tracked impact velocity, continuous foot--ground contact), as is merging
the non-uniform-wear and multi-joint analyses above into a single
profile-evolving coupled loop.

\section{Proposed experimental validation}

This study is computational, so we specify a concrete, falsifiable validation
protocol rather than claim measured data. A single Jansen leg is instrumented at
the load-bearing crank--ground pin ($G$:$c$--ground)---the joint identified here
as dominating wear. A hardened-steel pin runs in a replaceable bored bushing whose
radial clearance $c_0$ is set by selective fit (nominal $\SI{100}{\micro\metre}$
and $\SI{200}{\micro\metre}$, matching the simulated cases); the crank is driven
at the modelled speed by a servo with a phase encoder, and the stance
ground-reaction is reproduced by a vertical load at the foot. Four model
predictions are then directly testable (Table~\ref{tab:validation}). (i)
\emph{Impact amplification}: a force/acceleration sensor at the instrumented joint
should record stance-phase contact-force peaks $\sim\Pamp\times$ the smooth
ideal-joint reaction; their absence would refute the impact mechanism. (ii)
\emph{Per-trajectory sensitivity}: repeating the run from controlled but varied
starting phases should reproduce a wide cycle-to-cycle scatter in peak force---a
smooth, perfectly repeatable signal would contradict the predicted impact
sensitivity. (iii) \emph{Statistical durability advantage}: accelerated wear tests
on matched Jansen and optimized legs, across several bushing specimens and
randomized phases, should show the optimized joint's worn clearance and scar
volume growing $\sim\WensA$--$\WensB\times$ slower \emph{in the mean}; because
single specimens are predicted to be chaotic, the comparison must be made over an
ensemble. (iv) \emph{Non-uniform wear scar}: post-test profilometry of the bore
should reveal wear concentrated on a narrow load-bearing arc---locally far deeper
than a uniform-wear assumption predicts (Section~5)---confirming the worn-profile
mechanism.

\begin{table}[htbp]
\centering
\caption{Falsifiable predictions and their experimental signatures}
\label{tab:validation}
\small
\begin{tabular}{p{0.26\linewidth}p{0.30\linewidth}p{0.30\linewidth}} \toprule Prediction & Measurement & Confirming signature\\
\midrule
Impact amplification & joint load cell / accelerometer & peak $\sim\Pamp\times$ ideal reaction\\
Per-trajectory chaos & peak force vs.\ starting phase & wide non-repeatable scatter\\
Durability advantage & worn clearance \& scar volume (ensemble) & optimized $\sim\WensA$--$\WensB\times$ slower in mean\\
Non-uniform scar & bore profilometry & wear localized on load arc\\
\bottomrule
\end{tabular}
\end{table}

\section{Conclusion}

We built the first clearance-coupled forward-dynamic model of the Jansen leg
and used it to probe how the mechanism degrades in service. Three robust messages
emerge. First, neglecting joint clearance underestimates peak loads by
$\sim\Pamp\times$, and letting two load-bearing joints carry clearance
simultaneously raises ensemble peak forces further (to $\sim\Fmultipk\,$N); a
clearance-free wear model thus understates load and wear severity. Second,
although the wear--clearance--impact coupling is impact-sensitive---single
trajectories are chaotic and can even reverse the design ranking---an ensemble
comparison shows the gait--durability-optimized design of the companion study to
be robustly more durable, with per-cycle wear $\sim\WensA$--$\WensB\times$ lower
(peak force $\sim\Fens\times$ lower) at one clearance joint and still
$\sim\Wmulti\times$ lower on both with two ($p<0.01$ throughout). Third, the wear is
strongly non-uniform---it concentrates on a $\sim\Warc^\circ$ load arc---so the
common uniform-clearance-growth assumption underestimates local clearance growth
by $\sim\Wconc\times$. The clearance-free durability advantage therefore survives
the chaotic, multi-joint, non-uniformly-worn coupling in the statistical mean,
provided it is judged over an ensemble. We give a falsifiable experimental
protocol (Section~6) for each prediction; merging the non-uniform-wear and
multi-joint analyses into a single profile-evolving coupled loop is the natural
next step the released model supports.

\bibliographystyle{unsrtnat}
\bibliography{refs}

\end{document}